%% file: main.tex
\documentclass[10pt]{article} 
\usepackage[preprint]{tmlr}

\input{math_commands.tex}

\usepackage{hyperref}
\usepackage{url}

\usepackage{glossaries-extra}
\glsdisablehyper
\usepackage{hyperref}
\usepackage{url}

\usepackage{graphicx}
\usepackage{algorithm}

\newacronym{cnn}{CNN}{convolutional neural network}
\newacronym{gru}{GRU}{gated recurrent unit}
\newacronym{lstm}{LSTM}{long short-term memory}
\newacronym{mdp}{MDP}{Markov decision process}
\newacronym{mse}{MSE}{mean squared error}
\newacronym{mlp}{MLP}{multi-layer perceptron}
\newacronym{rl}{RL}{reinforcement learning}
\newacronym{rnn}{RNN}{recurrent neural network}
\newacronym{nlp}{NLP}{natural language processing}
\newacronym{cv}{CV}{computer vision}
\newacronym{pomdp}{POMDP}{partially observed Markov decision process}
\newacronym{marl}{MARL}{multi-agent reinforcement learning}
\newacronym{vit}{ViT}{vision transformer} 
\newacronym{iid}{i.i.d}{independent and identically distributed} 
\newacronym{medt}{MeDT}{medical decision transformer}
\newacronym{dt}{DT}{decision transformer}
\newacronym{wis}{WIS}{weighted importance sampling}
\newacronym{mimic}{MIMIC-III}{medical information mart for intensive care}
\newacronym{icu}{ICU}{intensive care unit}
\newacronym{ehr}{EHR}{electronic health record}
\newacronym{gpt}{GPT}{generative pre-trained transformer}
\newacronym{iv}{IV}{intravenous fluid}
\newacronym{vp}{VP}{vasopressor}
\newacronym{ope}{OPE}{off-policy evaluation}
\newacronym{ddpg}{DDPG}{deep deterministic policy gradient}
\newacronym{atg}{ATG}{acuity-to-go}
\newacronym{rtg}{RTG}{returns-to-go}
\newacronym{bc}{BC}{behaviour cloning}
\newacronym{fqe}{FQE}{fitted Q-evaluation}
\newacronym{wdr}{WDR}{weighted doubly robust}
\newacronym{bcq}{BCQ}{batch constrained Q-learning}
\newacronym{is}{IS}{importance sampling}
\newacronym{am}{AM}{approximate model}
\newacronym{nfqi}{NFQI}{neural fitted Q-learning}
\newacronym{cql}{CQL}{conservative Q-learning}
\newacronym{ddqn}{DDQN}{double deep Q-learning}

\newcounter{daggerfootnote}

\title{Empowering Clinicians with Medical Decision Transformers: A Framework for Sepsis Treatment}

\author{\name Aamer Abdul Rahman$^{1,3}$, Pranav Agarwal$^{3}$, Rita Noumeir$^{3}$, Philippe Jouvet$^{2,4}$, \\ Vincent Michalski\thanks{Equal senior author contribution.} $^{,1,2}$, Samira Ebrahimi Kahou$^{\ast,1,5,6}$ \\
      \addr  $^1$Mila \quad $^2$Université de Montréal \quad $^3$École de Technologie Supérieure \quad $^4$CHU Sainte-Justine Hospital \\ $^5$Canada CIFAR AI Chair \quad $^6$ University of Calgary\\
      \texttt{ar.aamer@gmail.com}}

\def\month{MM}  
\def\year{YYYY} 
\def\openreview{\url{https://openreview.net/forum?id=XXXX}} 

\begin{document}

\maketitle

\begin{abstract}
Offline reinforcement learning has shown promise for solving tasks in safety-critical settings, such as clinical decision support. Its application, however, has been limited by the lack of interpretability and interactivity for clinicians. To address these challenges, we propose the \emph{medical decision transformer (MeDT)}, a novel and versatile framework based on the goal-conditioned reinforcement learning paradigm for sepsis treatment recommendation. \emph{MeDT} uses the decision transformer architecture to learn a policy for drug dosage recommendation. During offline training, \emph{MeDT} utilizes collected treatment trajectories to predict administered treatments for each time step, incorporating known treatment outcomes, target acuity scores, past treatment decisions, and current and past medical states. This analysis enables \emph{MeDT} to capture complex dependencies among a patient's medical history, treatment decisions, outcomes, and short-term effects on stability. Our proposed conditioning uses acuity scores to address sparse reward issues and to facilitate clinician-model interactions, enhancing decision-making. Following training, \emph{MeDT} can generate tailored treatment recommendations by conditioning on the desired positive outcome (survival) and user-specified short-term stability improvements. We carry out rigorous experiments on data from the MIMIC-III dataset and use off-policy evaluation to demonstrate that \emph{MeDT} recommends interventions that outperform or are competitive with existing offline reinforcement learning methods while enabling a more interpretable, personalized and clinician-directed approach.
\end{abstract}

\section{Introduction}
\label{sec:intro}
Sepsis is a fatal medical condition caused by the body's extreme response to an infection. Due to the rapid progression of this disease, clinicians often face challenges in choosing optimal medication dosages. Hence, there is significant interest in developing clinical decision support systems that can help healthcare professionals in making more informed decisions~\citep{sutton2020overview}. In the medical field, many tasks involve sequential decision-making, such as evaluating a patient's evolving condition in the \gls{icu} to make informed medical interventions. This is where \gls{rl} comes in as a promising solution for developing policies that recommend optimal treatment strategies for septic patients~\citep{raghu2017deep,Komorowski2018TheAI,Killian2020AnES,Saria2018IndividualizedST,huang22continuous}.

These tools are intended to bolster and assist healthcare workers rather than replace them \citep{gottesman2018evaluating}. Therefore, the reward function employed by these \gls{rl} algorithms ideally necessitates clinician input to ensure that the policy generates decisions aligned with the domain expert's intentions \citep{gottesman2019guidelines}. However, the majority of existing studies predominantly depend on binary reward functions, signifying the patient's mortality \citep{Komorowski2018TheAI, Killian2020AnES,tang2022leveraging}. In other words, the reward at each timestep in the patient's history remains zero until the final interval of the episode. This design leaves no room for clinician input to modulate the policy toward the achievement of desirable tasks, such as the stabilization of certain vital signs.

Existing works~\citep{Killian2020AnES,Lu2020IsDR,Li2019OptimizingSM} often rely on modeling the patient's medical history using \glspl{rnn}. These networks struggle with complex and long medical records due to vanishing or exploding gradients~\citep{pascanu2013difficulty}, leading to sub-optimal \gls{rl} policies~\citep{parisotto2020stabilizing}. Sparse rewards also pose challenges in the learning of optimal policies since it can be difficult to identify a causal relationship between an action and a distant reward~\citep{sutton2018reinforcement}. The sequential design of \glspl{rnn} aggravates this problem. 
The low interpretability of model reasoning is another problem, given the high-stakes nature of clinical decision making. It is essential to address these challenges to create reliable decision support systems and improve clinical uptake of machine learning solutions.
Transformers~\citep{vaswani2017attention} are shown to effectively model long sequences, which enables learning of better representations for treatment histories of patients,  potentially yielding more informed predictions. 

In this paper, we propose the \emph{\gls{medt}}, an offline \gls{rl} framework where treatment dosage recommendation for sepsis is framed as a sequence modeling problem. \emph{\gls{medt}}, as shown in Figure~\ref{fig:MedDT}, is based on the \gls{dt} architecture~\citep{chen2021decision}. It recommends optimal treatment dosages by autoregressively modeling a patient's state while conditioning on hindsight returns. To provide the policy with more informative and goal-directed input, we also condition \emph{\gls{medt}} on one-step look-ahead patient acuity scores~\citep{leGall1993ANS} at every time-step. This enhances the potential for more granular conditioning while facilitating the interaction of domain experts with the model.

Below we summarize the main contributions of this work:
\begin{itemize}
    \item We propose \emph{\gls{medt}}, a transformer-based policy network that models the full context of a patient's clinical history and recommends medication dosages.
    \item We develop a framework to enable clinicians to guide the generation of treatment decisions by specifying short-term target improvements in patient stability, which addresses the sparse reward issue. 
    \item We demonstrate that \emph{\gls{medt}} outperforms or is competitive with popularly used offline \gls{rl} baselines over multiple methods of \gls{ope} such as \gls{fqe}, \gls{wdr} and \gls{wis}. Additionally, we leverage a transformer network, the \emph{state predictor}, to serve as an approximate model to capture the evolution of a patient's clinical state in response to treatment. This model enables autoregressive inference of \emph{\gls{medt}} and also serves as an interpretable evaluation framework of models used for clinical dosage recommendation.
\end{itemize}

\section{Related Work}
\label{sec:related}

\subsection{RL for Sepsis Treatment} 
\label{subsec:rlsepsis}
The use of \gls{rl} in sepsis treatment aims to deliver personalized, real-time decision support. It involves modeling optimal strategies for the administration of treatments, such as \glspl{vp} and \glspl{iv}, based on patient data and expert advice. This problem poses a considerable challenge due to the potential for long-term effects associated with these treatments, such as the accumulation of interstitial fluid and subsequent organ dysfunction resulting from excessive fluid administration~\citep{gottesman2018evaluating}.

To address this issue, \citet{Komorowski2018TheAI} propose a value-iteration algorithm using discretized patient data from \glspl{ehr} for treatment action selection. Subsequent work uses Q-learning with continuous states and discrete actions and employs \gls{ope} to evaluate policies~\citep{raghu2017deep}. 
\citet{huang22continuous} uses \gls{ddpg} with continuous states and actions to provide precise dosage recommendations. Other works explore model-based \gls{rl}~\citep{Peng2018ImprovingST} and combined deep \gls{rl} with kernel-based \gls{rl}~\citep{Raghu2018ModelBasedRL} to further improve treatment recommendations for septic patients.
Yet, several significant issues still need to be resolved, which currently impede the practical implementation of \gls{rl} for the treatment of sepsis.
Most of these studies assume that agents begin with a baseline reward of zero until the end of treatment. At the final time-step in a patient's history, a positive reward is given for survival and a negative reward otherwise. 
Since the manifestation of treatment outcomes (mortality) can occur with a delay of several days after decisions are made, it is challenging to identify effective treatment strategies. 
Shorter-term objectives, such as the stabilization of vital signs, are often overlooked. 
Additionally, given the wealth of data being generated for each \gls{icu} patient, identifying the most relevant aspects in the treatment history may not be immediately apparent~\citep{gottesman2018evaluating}.

\subsection{Transformer-based Policies}
\label{subsec:rltransformer}
Another challenge in treatment modeling is introduced by the partial observability of the patient's state at each time-step. A single reading of vital signs provides incomplete information on the patient's well-being. \Glspl{rnn} address this issue by sequentially processing multiple time-steps of data, but face difficulties in capturing a patient's complete state history due to unstable gradients~\citep{pascanu2013difficulty}. This may result in incomplete information and, consequently, inaccurate decision-making~\citep{yu2021reinforcement}. Recent research in \gls{rl}~\citep{parisotto2020stabilizing,parisotto2021efficient,janner2021offline,tao2022evaluating} is shifting towards attention-based networks~\citep{niu2021review} like transformers~\citep{vaswani2017attention,lin2022survey}, which process information from past time-steps in parallel.

Transformers better capture long contexts and can be effectively trained in parallel~\citep{lin2022survey}. This addresses challenges posed by the sequential processing in \glspl{rnn}~\citep{wen2022transformers}. The self-attention mechanism in transformers is particularly beneficial, addressing issues related to sparse or distracting rewards. 
Self-attention, in short, first computes attention weights for information in each time-step by matching their corresponding \emph{keys} and \emph{queries}, which are learnable projections of input tokens. Afterwards, these weights are used to compute weighted sums of \emph{values} corresponding to each time-step, potentially discovering dependencies between distant time-steps.
\Gls{dt}~\citep{chen2021decision} leverages these advantages for offline \gls{rl}~\citep{furuta2021generalized,xu2022prompting,meng2021offline}, by conditioning a policy on the full history of states, actions and an observed or desired \emph{reward-to-go}.
Building on the \gls{dt} architecture, we propose \emph{\gls{medt}}, which integrates additional conditioning via short-term goals for improvements in patient vital signs, yielding a framework for effective sepsis treatment recommendation.

\begin{figure*}[t!]
    \centering
    \includegraphics[width=0.95\linewidth]{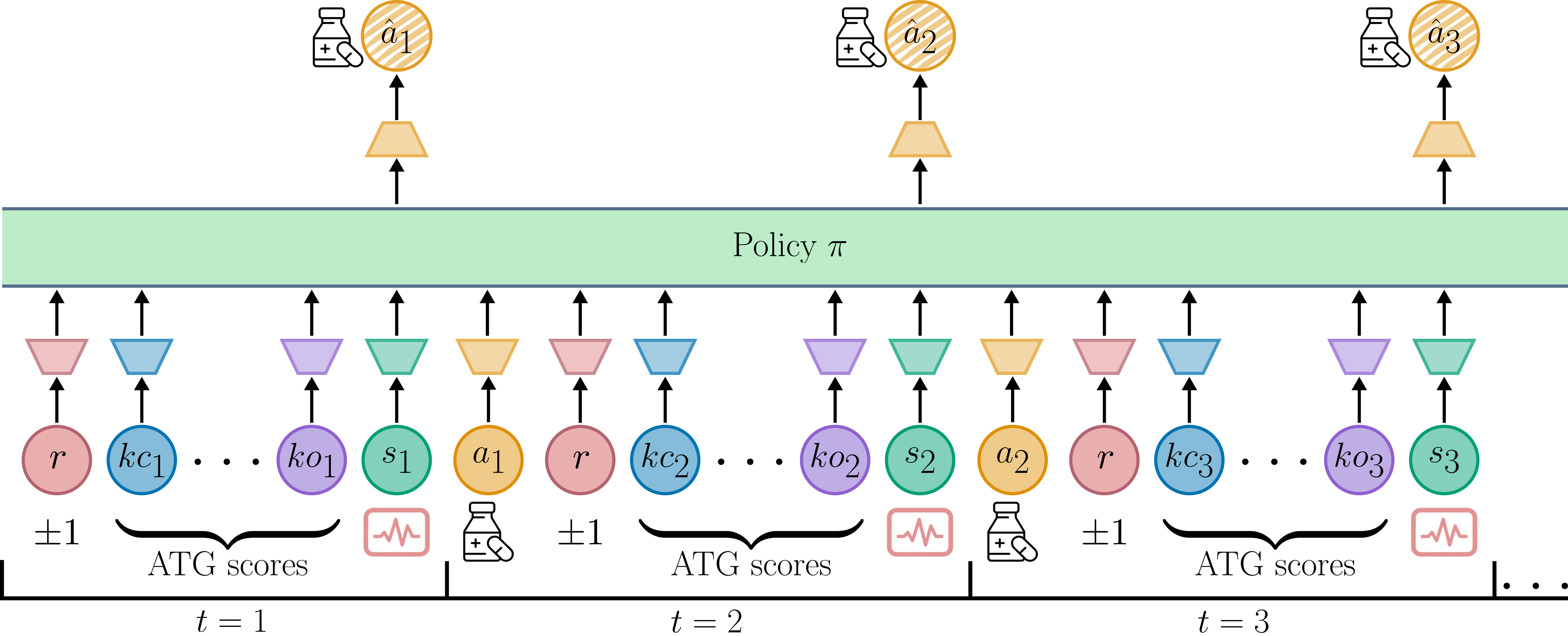}
    \caption{ \emph{MeDT} training: At each time-step $t$, the \emph{MeDT} policy attends to the past treatment trajectory. This includes the desired treatment outcome $r$ (at inference time fixed to $+1$ indicating survival), desired next-step acuity scores $k_1,\dots,k_t$ where $k_t = (kc_t, kr_t, kn_t, kl_t, kh_t, km_t, ko_t)$, patient states $s_1, \dots, s_t$, administered drug doses $a_1, \dots, a_{t-1}$, and outputs a dose prediction $\hat{a}_t$.
    }
    \label{fig:MedDT}
\end{figure*}

\begin{figure*}[t!]
    \centering    \includegraphics[width=0.9\linewidth]{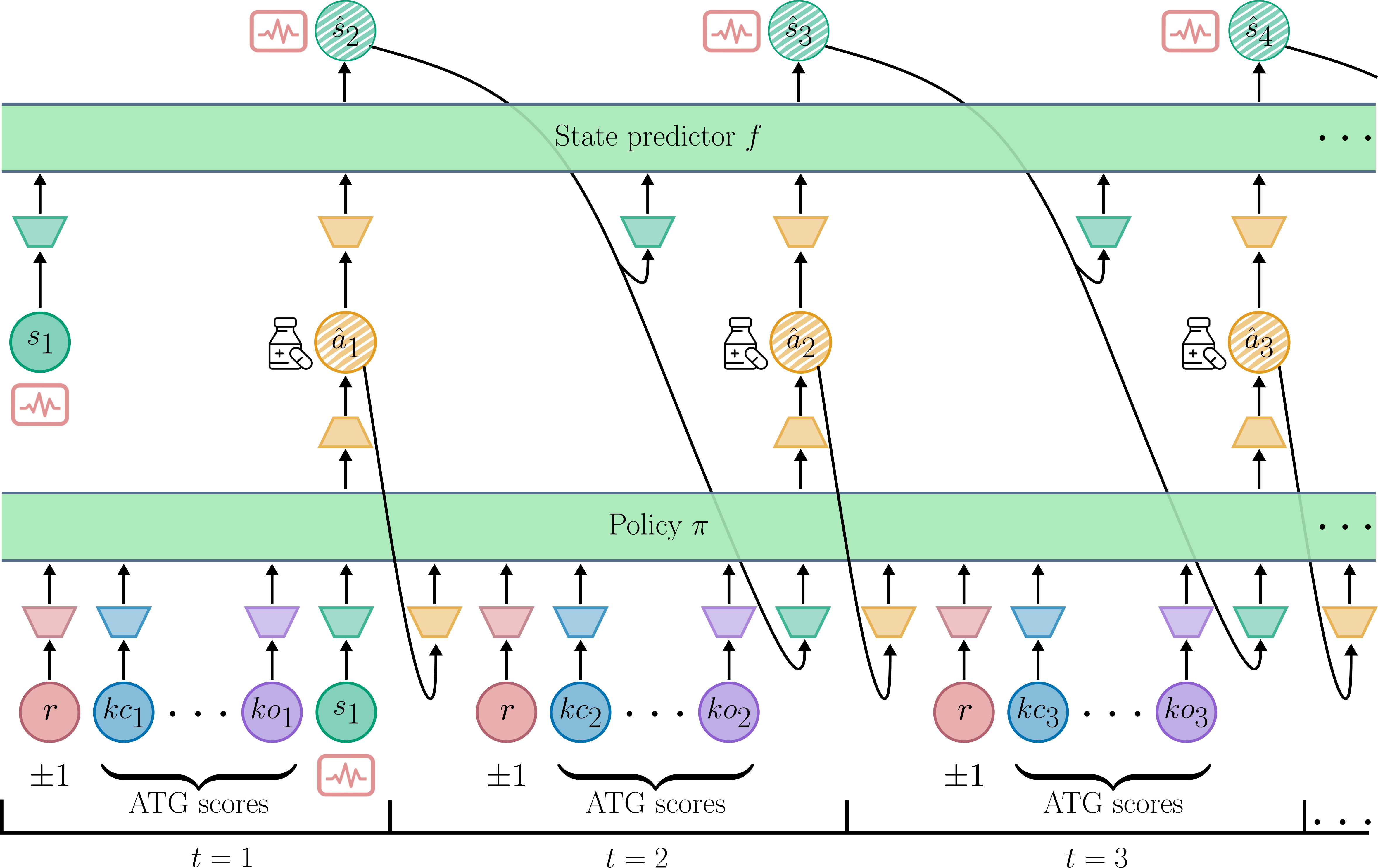}
    
    \caption{Autoregressive evaluation pipeline: At each time-step $t$, the pre-trained state predictor attends to past recommended doses $\hat{a}_1, \dots, \hat{a}_{t}$, the initial patient state $s_1$ and predicted patient states $\hat{s}_2, \dots, \hat{s}_t$, and outputs a prediction  $\hat{s}_{t+1}$ of the patient state at time $t+1$. Both dosage recommendations $\hat{a}_{t+1}$ and predicted states are fed back to \emph{MeDT} to simulate treatment trajectories with multiple sequential decisions.}
    \label{fig:medt_inference}
\end{figure*}

\subsection{Off-Policy Evaluation}
\label{subsec:relatedope}
\Gls{ope} is a fundamental problem in \gls{rl} concerned with estimating the expected return of a given decision policy using historical data obtained by different behavior policies~\citep{uehara2022review}. Such an evaluation strategy is particularly useful in situations where interacting with the environment is costly, risky, or ethically challenging, like in healthcare~\citep{sutton2018reinforcement, precup2000eligibility,gottesman2020interpretable,sheth2022learning}. However, \gls{ope} is inherently difficult because it necessitates counterfactual reasoning, i.e. unraveling what would have occurred if the agent had acted differently based on historical data. Nevertheless, while \gls{ope} may not necessarily help learn the optimal policy, it can help identify policies with lower suboptimality~\citep{tang2021model}.

A large subset of \gls{ope} methods are based on the concept of \gls{is}. \Gls{is} uses validation data to assess the evaluation policy's value by adjusting the weight of each episode based on its relative likelihood~\citep{puaduraru2013empirical, voloshin2019empirical}. The \gls{wis} estimator is considered more stable than \gls{is}~\citep{puaduraru2013empirical, voloshin2019empirical}. On the other hand, we can directly estimate the evaluation policy's performance using the Q-function with \gls{fqe}, rather than adjusting the weights of observed experiences like \gls{wis}~\citep{le2019batch}. \Gls{fqe} predicts the expected cumulative reward for taking a specific action in a given state. \Gls{wdr} is another \gls{ope} technique that combines two approaches for estimating policy value~\citep{thomas2016data,jiang2016doubly}. It leverages importance sampling, which adjusts the weight of past experiences based on their likelihood. Additionally, WDR incorporates value estimates at each step to improve accuracy and reduce overall variation in the learning process. Finally, \glspl{am} are another class of \gls{ope} that involves directly modeling the dynamics of the environment~\citep{jiang2016doubly,voloshin2019empirical}. This approximation, while not exact, may be sufficient to evaluate the policy's performance. \citet{tang2021model} empirically demonstrated that \gls{fqe} provided the most accurate and stable estimations over varying data conditions. Given the difficult nature of evaluation in this problem setting, we utilize each of the four mentioned \gls{ope} methods to infer rigorous and robust policy evaluations.

\subsection{Interpretability}
\label{subsec:relatedinterp}
The need for interpretability is more significant in safety-critical fields such as healthcare~\citep{amann2020explainability}. Despite extensive research, the deployment of deep learning in healthcare has been met with resistance \citep{yin2021role}. This is primarily due to the black-box nature of these networks, resulting from their complexity and large number of parameters. Moreover, attaining interpretability in \gls{rl} has been a major challenge hindering its deployment \citep{agarwal2023transformers}. Owing to their complexity, existing \gls{rl} algorithms fall short of being fully interpretable~\citep{glanois2021survey}.

One simple method of attempting to understand the inner workings of transformers, is to visualize the computed attention weights. However, \citet{serrano-smith-2019-attention} show that attention weights only produce noisy predictions of the relevance of each input token.
Recent works delve into formulating methods that more representatively capture the relevance of input tokens. \citet{abnar2020quantifying} propose the rollout method, which considers paths over the pairwise attention graph while assuming that attention is computed linearly. However, this work is shown to assign importance to irrelevant tokens~\citep{chefer2021transformer}. \citet{chefer2021transformer} introduce a method based on layer-wise relevance propagation, which is effective for encoder transformers.
\citet{chefer2021generic} present a generic interpretability method that is compatible with every type of transformer architecture. The proposed method relies on the concept of information flow. It involves monitoring the mixing and evolution of attention to generate representative heatmaps, illustrating the importance assigned to input tokens in the model's decision. This method produces interpretations that are similarly or more accurate than prior methods while being simpler to implement. In this work, we utilize this method to generate interpretations of \emph{\gls{medt}} and visualize the relevance assigned in the input space to aid clinicians in understanding the rationale behind the model's decision-making.
\begin{figure*}[t]
    \centering
    \includegraphics[width=0.98\linewidth]{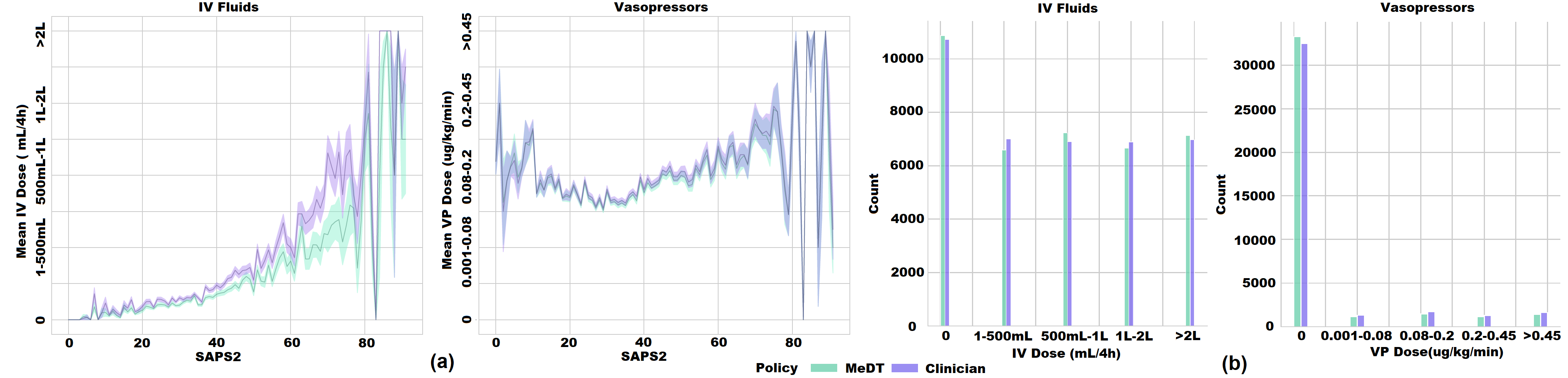}
    \caption{(a) Dosage recommended by \emph{\gls{medt}} and clinician policy for different SAPS2 scores. (b) Distribution of \gls{iv} fluids and \glspl{vp} given by the \emph{\gls{medt}} and clinician policies.
}
    \label{fig:iv_vp_disbn} 
\end{figure*}
\section{Medical Decision Transformer (MeDT)}
\label{sec:medt}
We frame our problem as a \acrfull{mdp}, comprising a tuple $\mathcal{(S,A,P,R,S')}$, where $\mathcal{S}$ denotes the set of possible patient states, $\mathcal{A}$ the set of possible dosage recommendations, $\mathcal{P}$ the state transition function, $\mathcal{R}$ the reward function and $\mathcal{S'}$ is the next patient state. 
While this framework is well-suited for learning policies via trial-and-error using \gls{rl} methods, direct interaction with the environment can be risky in safety-critical applications like clinical decision making. To mitigate this risk, we use offline \gls{rl}, a subcategory of \gls{rl} that learns an optimal policy using a fixed dataset of collected trajectories each containing the selected actions, observed states and obtained rewards.

\Gls{dt}~\citep{chen2021decision} uses transformers to model offline \gls{rl} via an upside-down \gls{rl} approach, where a policy has to select actions, that are likely to yield a specified future reward for a given past trajectory~\citep{schmidhuber2019reinforcement}. Our proposed \emph{\gls{medt}} architecture follows a similar approach for learning policies.

The input tokens for the policy model encode past treatment decisions and patient states, as well as the desired \gls{rtg}. The output at each time-step is a distribution over possible actions. Specifically, we condition the model using \gls{rtg} $r_t=\sum_{t^{\prime}=t}^T R_{t^{\prime}}=R_T$, which represents the singular positive or negative treatment outcome at the last time-step, similar to \gls{dt}. In addition, we propose to condition \emph{\gls{medt}} on short-term goals, such as future patient acuity scores, or \gls{atg}. The acuity score provides an indication of the severity of the illness of the patient in the \gls{icu} based on the status of the patient's physiological systems and can be inferred from vital signs of the corresponding time-step. Higher acuity scores indicate a higher severity of illness. In this work, we opt to use the SAPS2~\citep{leGall1993ANS} acuity score as opposed to popular scores such as SOFA~\citep{jones2009sequential}. This is because SAPS2 considers relatively more physiological variables, which we believe will provide more informative conditioning, allowing flexible user interactions~\citep{morkar2022comparative}. This formulation allows clinicians to input desired acuity scores for the next state, providing additional context for treatment selection. This leads to more information-dense conditioning, allowing clinicians to interact with the model and guide the policy's generation of treatment dosages.

To enable clinicians to provide more detailed inputs, we break down the SAPS2 score into constituent scores that correspond to specific organ systems~\citep{leGall1993ANS}. Following the definitions provided by~\citet{Schlapbach22}, we define split scores $k$ = ($kc$, $kr$, $kn$, $kl$, $kh$, $km$, $ko$) to represent the status of the cardiovascular, respiratory, neurological, renal, hepatic, haematologic and other systems, respectively. A more detailed breakdown of these scores can be found in the Appendix in Table~\ref{table:saps}.

In addition to the specification of the treatment outcome, our overall framework empowers domain experts to establish and select a scheme for interpretable short-term goal-conditioning, 
allowing clinicians to guide the model using their knowledge of the relation between intermediate goals, such as maintaining patients' vital signs within a specified range, and favorable long-term outcomes such as reduced mortality. This enhances the usability of the model for clinicians, enabling efficient interaction with the model for future dosage recommendations, considering the current state of the patient's condition. 
It is important to note that defining short-term goals presents a challenge, given the ongoing complexity in determining ideal targets for sepsis resuscitation \citep{simpson2017septic}. Using these scores, the treatment progress over $T$ time steps forms a trajectory

\begin{equation}
\tau=\left((r_1, k_1, s_1, a_1), (r_2, k_2, s_2, a_2), \ldots, (r_T, k_T, s_T, a_T)\right),
\label{eq:trajec}
\end{equation}
where, for each time-step $t$, $r_t, k_t, s_t, a_t$ respectively correspond to the \gls{rtg}, the \gls{atg}, the patient state, and the treatment decisions.
We train our policy, a causal transformer network, to predict ground truth dosages that were administered by a clinician at each time-step given the treatment trajectory, ignoring future information and the prediction target via masking (Figure \ref{fig:MedDT}). \emph{\Gls{medt}} aims to learn an optimal policy distribution 
\begin{equation}
P_{\pi}(a_{t}|s_{\leq t},r_{\leq t},k_{\leq t},a_{<t}), 
\label{eq:policy_disbn}
\end{equation}
inspired by the model architecture used in \citet{chen2021decision}. We use an encoder with a linear layer and a normalization layer for each type of input (i.e. \gls{rtg}, \gls{atg}, state, action) to project raw inputs into token embeddings. To capture the temporal dynamics of the patient's trajectory, we use learned position embeddings for each time-step, which are added to the token embeddings. Finally, the resultant embeddings are fed into a causal transformer, which autoregressively predicts the next action tokens.
\subsection{Evaluation} 
In online \gls{rl}, policies are typically assessed by having them interact with the environment. However, healthcare involves patients, and employing this evaluation method is unsafe. In this work, we evaluate the learned \gls{rl} policy in an observational setting, where the treatment strategy is assessed based on historical data \citep{gottesman2018evaluating}. Following the model-based \gls{ope} approach, we introduce an additional predictor network based on the causal transformer~\citep{radford2018improving}.

The predictor network, shown in Figure \ref{fig:medt_inference}, acts as a stand-in for the simulator during inference. It is trained to learn a state-prediction model defined by the distribution 
\begin{equation}
P_{\theta}(s_{t}|a_{< t},s_{< t}), 
\label{eq:state_disbn}
\end{equation}
using a similar architecture as the policy network. This allows us to model how a patient's state changes in response to medical interventions. Rather than introducing a termination model, we use a fixed rollout length of $\mathcal{H}$. The estimated acuity scores can provide more clinically relevant estimates because they indicate how the stability of the physiological state of the patient may change given a treatment policy. While not exact, this approximation can prove adequate for generating reasonable estimates of a patient's physiological dynamics. This enables inferring estimates of patient acuity scores (SAPS2) from predicted states, which can then be used for policy evaluation. Furthermore, during inference, this model allows autoregressive generation of a sequence of actions by predicting how the patient's state evolves as a result of those actions. Figure~\ref{fig:medt_inference} and Algorithm~\ref{alg:evalloop} depict this rollout procedure.

\begin{algorithm}[h!]
  \caption{Evaluation Loop}%
  \label{alg:evalloop}%
{%
\begin{enumerate}
  \item \textbf{Input:} Initial patient state $s_0$
  \item \textbf{Output:} Acuity score $g_1, \ldots, g_T$
  \item Set target return $r_T = 1$
  \item Initialize state $s_1 = s_0$, target return $r_{1:T} = r_T$ and action sequence $a_{0} = \{\}$
  \item \textbf{for} {$t=1,2,\ldots,T$}
  \item Select desired Acuity To Go $k_t$
  \item Select action $a_t = \mathrm{MeDT}(r_{1:t},k_{1:t},s_{1:t},a_{0:t-1})$
  \item Append $a_t$ to the sequence of actions: $a_{1:t}=a_{0:t-1}+[a_t]$
  \item Estimate new state: $s_{t+1} = \mathrm{state\_estimator}(s_{1:t},a_{1:t})$
  \item Evaluate acuity score $g_{t+1}$ for state $s_{t+1}$
  \item Append $s_t$ to the sequence of states: $s_{1:t}=s_{1:t-1}+[s_t]$
  \item \textbf{end for}
  \item \textbf{return} acuity score $g_1,\ldots ,g_T$
\end{enumerate}
}%
\end{algorithm}

Additionally, we use \gls{wis}~\citep{puaduraru2013empirical, voloshin2019empirical}, \gls{wdr}~\citep{jiang2016doubly,voloshin2019empirical} and \gls{fqe}~\citep{le2019batch} to rigorously evaluate the performance of policies.

\Gls{wis} uses a behavior policy $\pi_{b}$ to evaluate a policy $\pi$ by re-weighting episodes according to their likelihood of occurrence~\citep{puaduraru2013empirical, voloshin2019empirical}. With the per-step importance ratio $\rho_t=\frac{\pi\left(a_t \mid s_t\right)}
{\pi_b\left(a_t \mid s_t\right)}$ and cumulative importance ratio $\rho_{1: t}=\prod_{t^{\prime}=1}^t \rho_{t^{\prime}}$, \gls{wis} can be computed as
\begin{equation}
\frac{1}{N} \sum_{n=1}^N \frac{\rho_{1: T^{(n)}}^{(n)}}{w_{T^{(n)}}}\left(\sum_{t=1}^{T^{(n)}} \gamma^{t-1} r_{t}^{(n)}\right),
\end{equation}
where $N$ is the total number of episodes, $T^{(n)}$ is the total number of time-steps for episode $n$, $\gamma$ is the discount factor and the average cumulative importance ratio $w_t=\frac{1}{N} \sum_{n=1}^M \rho_{1: t}^{(n)}$. 

\Gls{fqe} is a value-based temporal difference algorithm that utilizes the Bellman equation to compute bootstrapped target transitions from collected trajectories and then uses function approximation to compute the $Q$ value of policy $\pi$. This can be formalized as
\begin{equation}
\frac{1}{N} \sum_{n=1}^N \sum_{a \in \mathcal{A}} \pi(a \mid s_{1}^{(n)}) \widehat{Q}_{\mathrm{FQE}}^\pi(s_{1}^{(n)}, a) ,
\end{equation}
where $\widehat{Q}_{\mathrm{FQE}}$ is the estimated $Q$ function of $\pi$.

\Gls{wdr} utilizes both value estimators from \gls{fqe} as well as importance sampling from \gls{wis} in order to reduce the overall variance of estimations~\citep{jiang2016doubly, thomas2016data}. The \gls{wdr} estimator is defined as follows:
\begin{equation}
\begin{aligned}
\frac{1}{N} \sum_{n=1}^N \sum_{t=1}^{T^{(n)}}\left[\frac{\rho_{1: t}^{(n)}}{w_t} \gamma^{t-1} r_t^{(n)}-\left(\frac{\rho_{1: t}^{(n)}}{w_t} \widehat{Q}^\pi\left(s_t^{(n)}, a_t^{(n)}\right)-\frac{\rho_{1: t-1}^{(n)}}{w_{t-1}} \widehat{V}^\pi\left(s_t^{(n)}\right)\right)\right] ,
\end{aligned}
\end{equation}
where $\widehat{Q}_{\mathrm{FQE}}$ and $\widehat{V}_{\mathrm{FQE}}$ is the estimated $Q$ and value function of policy $\pi$ respectively.

\begin{table*}[t]
  \caption{Estimated final patient acuity scores (averaged over 2,898 patients) for BCQ: batch constrained Q-learning, NFQI: Neural Fitted Q-Learning, DDQN: Double Deep Q-Learning, CQL: Conservative Q-Learning, BC: behavior cloning, DT: decision transformer and \emph{MeDT}: medical decision transformer.}
  \label{table1}
  \centering
  \begin{tabular}{|c|c|c|c|c|}
        \hline
        Models  & Overall $\downarrow$ & Low $\downarrow$ & Mid $\downarrow$ & High $\downarrow$ \\ 
        \hline
        BCQ & 42.10$\pm$0.03 & 41.78$\pm$0.07 & 42.38$\pm$0.03 & 41.59$\pm$0.15 \\
        NFQI & 44.21$\pm$0.05 & 43.26$\pm$0.08 & 44.85$\pm$0.06 & 45.13$\pm$0.13 \\
        DDQN & 43.47$\pm$0.04 & 43.08$\pm$0.07 & 43.64$\pm$0.05 & 43.29$\pm$0.11 \\
        CQL & 40.42$\pm$0.03 & 40.18$\pm$0.07 & 40.59$\pm$0.04 & 41.03$\pm$0.11 \\
        BC & 40.50$\pm$0.03 & 40.33$\pm$0.07 & 40.56$\pm$0.03 & 40.29$\pm$0.12 \\ 
        DT & 40.38$\pm$0.03 & 40.16$\pm$0.06 & 40.49$\pm$0.03 & \textbf{40.06}$\pm$\textbf{0.12}\\ 
        MeDT & \textbf{40.31}$\pm$\textbf{0.03} & \textbf{40.05}$\pm$\textbf{0.06} & \textbf{40.40}$\pm$\textbf{0.03} & 40.35$\pm$0.14 \\
        \hline
  \end{tabular}
  \label{tab:main_table}
\end{table*}
\subsection{Interpretability}
\label{subsec:methinterp}
We utilize the transformer interpretability method introduced by~\citet{chefer2021generic} which is based on the principle of information flow. We adapt this algorithm for the decoder transformer architecture of \emph{\gls{medt}} used in this work. This subsection outlines the mechanisms underlying the computation of relevance scores used to visualize interpretations.

Let $i$ refer to the input tokens of \emph{\gls{medt}}. $\mathbf{A}^{i i}$ represents the self-attention interactions between these tokens. Based on these interactions, we seek to compute the relevancy map $\mathbf{R}^{i i}$. Relevancy maps are constructed with a forward pass through the self-attention layers, where these layers attribute to aggregated relevance maps via the following propagation rules.

Given that each token is self-contained prior to attention operations, self-attention interactions are initialized with identity matrices. Thus, the relevancy maps are also initialized as identity matrices:
\begin{equation}
\mathbf{R}^{i i}=\mathbb{1}^{i \times i}.
\end{equation}
The attention matrix $\mathbf{A}$ from each layer is used to update the relevance maps. The gradients $\nabla \mathbf{A}$ are used to average over the heads $h$ dimension of the attention map, to account for the differing importance assigned across the heads of the matrix~\citep{voita2019analyzing}. $\nabla \mathbf{A}:=\frac{\partial y}{\partial \mathbf{A}}$, where $y$ refers to the output for which we wish to visualize relevance. The aggregated attention is then defined as:
\begin{equation}
\bar{\mathbf{A}}=\mathbb{E}_h\left((\nabla \mathbf{A} \odot \mathbf{A})^{+}\right),
\end{equation}
where $\mathbb{E}_h$ is the mean over the $h$ dimension and $\odot$ is the Hadamard product. $^+$ denotes that the negative values are replaced by zero prior to computing the expectation.

At each attention layer, these aggregated attention scores are then used to calculate the aggregated relevancy scores as follows:
\begin{equation}
\mathbf{R}^{s s}=\mathbf{R}^{s s}+\bar{\mathbf{A}} \cdot \mathbf{R}^{s s}.
\end{equation}
These relevancy scores can then be used to visualize the importance assigned across the input token space in the form of a heatmap. Since future tokens are masked in transformer decoders, there is more attention toward initial tokens in the input sequence. Hence, to apply these methods to \emph{\gls{medt}}, we normalize based on the receptive field of attention.
\begin{figure*}[t!]
    \centering
    \includegraphics[width=0.98\linewidth]{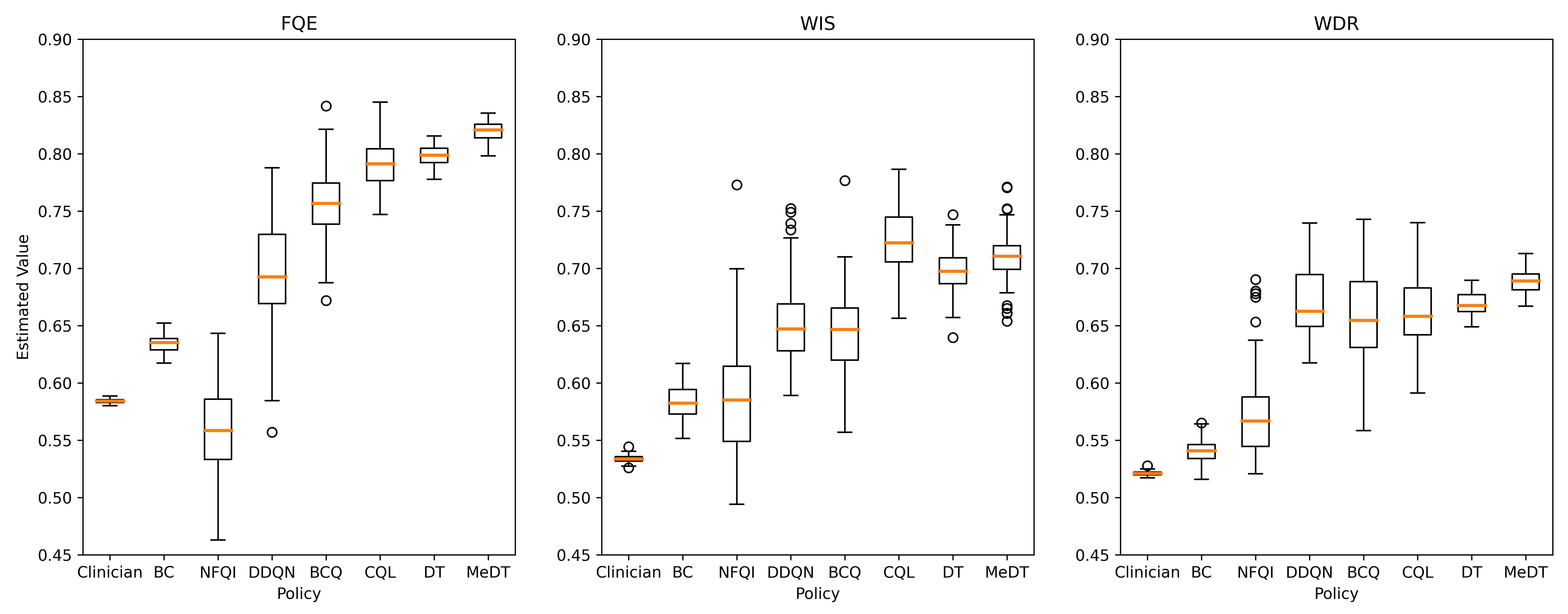}
    \caption{Box-plots of FQE, WIS and WDR off-policy evaluations for MeDT and baselines.
}
\label{fig:OPE}
\end{figure*}

\section{Experiments}
\subsection{Experimental Settings}
In this work, we train and evaluate the performance of \emph{\gls{medt}} on a cohort of septic patients. The cohort data is obtained from the \gls{mimic} dataset~\citep{johnson2016mimic}, which includes 19,633 patients, with a mortality rate of 9\%. These patients were selected on fulfilling the sepsis-3 definition criteria \citep{singer2016third}. To pre-process the data, we follow the pipeline defined by~\citet{Killian2020AnES}. We extract physiological measurements of patients recorded over 4-hour intervals and impute missing values using the K-nearest neighbor algorithm. Multiple observations within each 4-hour window are averaged.
 
The patient state consists of 5 demographic variables and 38 time-varying continuous variables such as lab measurements and vital signs. This work centers on the timing and optimal dosage of administering \gls{vp} and \gls{iv} fluids. The administration of each drug for patients is sampled at 4-hour intervals. We discretized the dosages for each drug into 5 bins, resulting in a combinatorial action space of 25 possible treatment administrations. Limiting our focus to \gls{iv} fluids and vasopressors implies that these are the only treatments within our control; other interventions like antibiotics that the patient might receive are outside the scope of our consideration.

\subsection{Baselines}
We compare \emph{\gls{medt}} to \gls{bcq}, \gls{nfqi}, \gls{ddqn} and \gls{cql} algorithms, which are commonly used baselines in recent works related to offline reinforcement learning~\citep{Killian2020AnES,tang2022leveraging,pace2023delphic}. Additionally, we train and evaluate \gls{dt} and a transformer-based \gls{bc} algorithm. \Gls{bc} refers to a transformer that takes as input past states and actions, guided by cross-entropy loss on predicted actions, to directly imitate the behavior of the clinician’s policy. \Gls{dt} builds on \gls{bc} by conditioning on returns-to-go. The proposed \emph{\gls{medt}} differs from \gls{dt} in that it also conditions on acuity-to-go at each time-step.

\subsection{Training}
The transformer policy is trained on mini-batches of fixed context length, which are randomly sampled from a dataset of offline patient trajectories. In our case, we chose a context length of 20, which is the longest patient trajectory in the dataset following pre-processing. For trajectories shorter than this length, we use zero padding to adjust them. During training, we use teacher-forcing, where the ground-truth sequence is provided as input to the model. At each time-step $t$, the \gls{atg} ($k_{t}$)  is set to the actual acuity scores of the state at time-step $t+1$ in the sequence. The prediction head of the policy model, associated with the input token $s_t$, is trained to predict the corresponding discrete treatment action $a_t$ using a cross-entropy loss. The loss for a complete trajectory is averaged over time-steps. Additionally, the state estimator is separately trained to predict the patient's state following the treatment actions. The prediction head of the state predictor model, corresponding to the input token $a_t$, is trained to estimate the continuous state $s_{t+1}$ using a mean square error loss. The models are trained on NVIDIA V100 GPUs. We aggregate experimental results for each model into mean and standard error over five random seeds. Additional details on hyperparameter selection can be found in the Appendix in Section~\ref{apd:hyperparameters}.

\begin{figure*}[t!]
    \centering
    \includegraphics[width=0.98\linewidth]{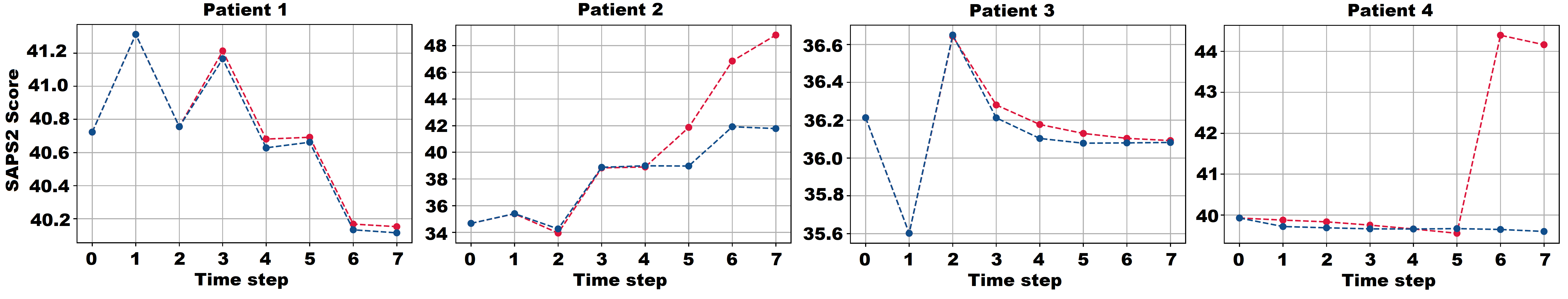}
       
    \caption{Visualization of 4 patient trajectories computed by the state predictor following treatment recommendation from \gls{dt} (red) and \emph{\gls{medt}} (blue).
}
\label{fig:iv_vp_traj}
\end{figure*}

\subsection{Results and Analysis}
We evaluate our proposed \emph{\gls{medt}} network in the autoregressive inference loop with the state predictor (Table \ref{tab:main_table}). As elaborated in the Appendix in Section~\ref{apd:data}, we use a naive heuristic to select \gls{atg}, and investigate whether the network conditioned on these prompts results in more stable patient outcomes. We compare our proposed approach to multiple baselines and run this loop over only 10 time-steps to avoid the accumulation of state-prediction errors resulting from the autoregressive nature of evaluation. We calculate the average and standard error of the SAPS2 scores of the states estimated by the predictor network for every patient in the test cohort. This cohort split comprises 2,945 patients. We also evaluate all policies over additional methods of \gls{ope} such as \gls{wis}, \gls{fqe} and \gls{wdr}.

\subsubsection{Quantitative Analysis}
From Table~\ref{table1}, we infer that the \emph{\gls{medt}} policy, which is conditioned on both positive \glspl{rtg} and our chosen \gls{atg} heuristic, results in the most stable estimated patient states. The \gls{dt} framework conditioned only with positive \glspl{rtg} performs better than \gls{bc} and other baselines. The learned policies are also evaluated for patients with different severity of sepsis (denoted as low, mid and high severity) based on the SAPS2 score of the initial state. Comparing the models, we observe that the \emph{\gls{medt}} policy results in more stable states for low and mid-severity patients, while \gls{dt} performs best for high-severity patients. We hypothesize that, given there are far fewer data samples for patients in the higher severity bracket, \emph{\gls{medt}} was not able to learn an accurate mapping of patient states to actions given the additional \gls{atg} context. This suggests that \emph{\gls{medt}} requires more samples relative to \gls{dt} to reach convergence.

We run an experiment to evaluate the sample efficiency of \gls{dt} and \emph{\gls{medt}} in Figure \ref{fig:scale} in the Appendix. We evaluate the performance of the policies when trained on 50\%, 75\% and 100\% of the data from the train split. \Gls{dt} performs better when trained on the smallest 50\% split, while \emph{\gls{medt}} is performant on the 75\% and 100\% splits. This supports the hypothesis that the additional conditioning used in \emph{\gls{medt}} has a negative impact on sample efficiency. It is worth noting that given the small size of the sepsis cohort from the MIMIC-III dataset, the 50\%, 75\% and 100\% splits are all low training sample settings relative to standard sizes of training data used in \gls{rl}. Nevertheless, this is an optimistic observation, given the potential for the exponential growth of data available on large-scale \glspl{ehr}.

Figure \ref{fig:OPE} depicts the results of the \gls{fqe}, \gls{wis} and \gls{wdr} evaluations. The \emph{\gls{medt}} policy produces the highest estimated values on \gls{fqe} and \gls{wdr} while \gls{cql} performs best on \gls{wis}. It is worth noting that the \emph{\gls{medt}} and \gls{dt} policies show noticeably less variance than the baselines, suggesting they are more robust models. These results indicate that the clinical dosage recommendations based on our proposed conditioning method may have had the intended treatment effects.

\begin{figure*}[t]
    \centering
    \includegraphics[width=0.98\linewidth]{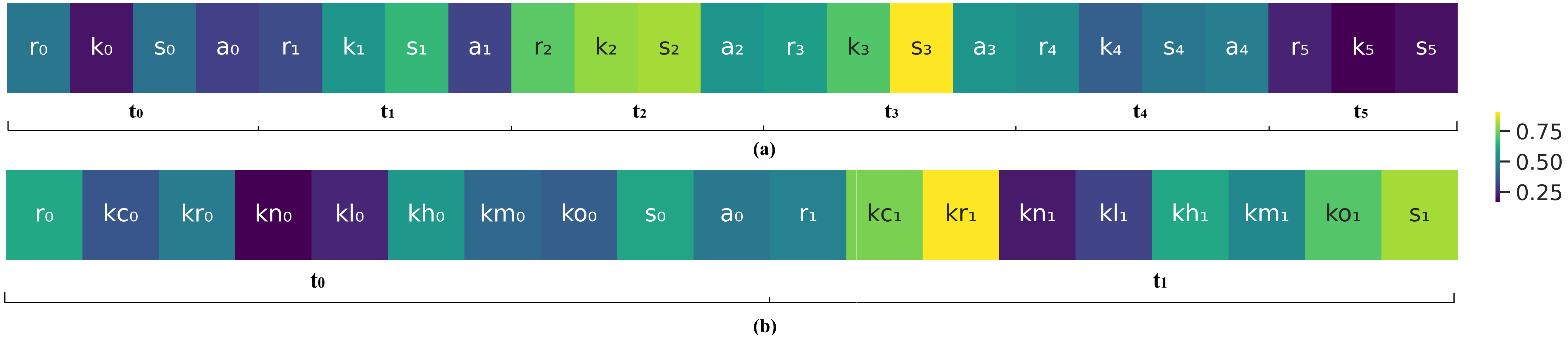} 
    \caption{Relevancy maps depicting the importance assigned by the model to input tokens upon predicting action $a_5$ for a sample patient trajectory. Darker and lighter colors indicate lower and higher relevance scores, respectively. In (a) the mean of the relevancy assigned to the \gls{atg} components is taken to better visualize the relevance across time. Heatmap (b) depicts relevance across each \gls{atg} component to visualize the importance assigned to each conditioning.
    }
    \label{fig:attention_map}
\end{figure*}

\subsubsection{Qualitative Analysis}
We qualitatively evaluate the policy of \emph{\gls{medt}} against the clinician's policy. To ensure accurate analysis, we use ground-truth trajectories as input sequences instead of relying on autoregressive inference, which may lead to compounding errors. In Figure~\ref{fig:iv_vp_disbn}a, we conduct a comparative analysis of the mean dose of \glspl{vp} and \glspl{iv} recommended by the \emph{\gls{medt}} policy and the clinician policy, for patient states with varying SAPS2 scores. Figure~\ref{fig:iv_vp_disbn}b presents the dosage distribution of \glspl{iv} and \glspl{vp} recommended by both the \emph{\gls{medt}} and clinician policies.

Our results show that the \emph{\gls{medt}} policy generally aligns with the clinician's treatment strategy but recommends lower doses of \glspl{iv} on average. Both policies exhibit a similar trend of increasing medication doses with worsening patient condition, for both \glspl{vp} and \glspl{iv}. Figure~\ref{fig:iv_vp_disbn}b reveals that the \emph{\gls{medt}} policy uses more zero dosage instances for both \glspl{iv} and \glspl{vp}, compared to the clinician policy. We hypothesize the significant overlap between the \emph{\gls{medt}} and clinician policy is a byproduct of the imbalanced nature of the dataset, given that over 91\% of patient trajectories in the dataset resulted in positive outcomes (survival). As a result, \emph{\gls{medt}} decides to imitate the clinician policy. Nevertheless, the alignment with the domain expert policy is ideal, especially in this high-stakes task where the algorithm relies solely on pre-existing static data for learning, as it is preferred to assess policies that only recommend subtle changes and closely resemble those of physicians as a precautionary measure~\citep{gottesman2019guidelines}.

Furthermore, previous studies have demonstrated a trend wherein lower dosages are recommended for patients with higher acuity scores~\citep{raghu2017deep}. This pattern can be linked to the common practice among clinicians of administering elevated dosages to individuals with high acuity scores, often associated with more severe medical conditions and, consequently, higher mortality rates. The challenge arises when algorithms lack data samples featuring high acuity scores coupled with minimal dosages. In such instances, these algorithms default to advocating lower medication doses. Figure~\ref{fig:iv_vp_disbn} demonstrates that the \emph{\gls{medt}} policy diverges from prior research by refraining from recommending minimal dosages for patients with elevated acuity scores. This serves as an indicator of better generalization and sample efficient properties from \emph{\gls{medt}} given this negative behavior is not observed.

In Figure~\ref{fig:iv_vp_traj}, we visualize the trajectories of multiple patients computed by the state predictor, following treatment actions recommended by both the \gls{dt} and \emph{\gls{medt}} policies. The impact of \gls{atg} conditioning on patient health is evident, as \emph{\gls{medt}} leads to more stable trajectories, demonstrating the potential of our framework to generate targeted and improved treatment recommendations by considering both the hindsight returns and \gls{atg} at each time-step. In the Appendix in Figure~\ref{fig:supp_traj}, we provide visualizations of some patient trajectories, where we observed that the \emph{\gls{medt}} policy produced the same or worse action policies relative to \gls{dt}, with no discernible effect of \gls{atg} conditioning. 
We hypothesize that this may be due to limitations of the dataset, which may not sufficiently cover some regions of the joint space of vital signs, treatment decisions and outcomes, causing the model to be unable to discover some causal relations.

\subsubsection{Interpretability} 
Currently, \gls{rl} algorithms typically function as opaque systems~\citep{gottesman2019guidelines}. They take in data and generate a policy as output, but these policies are often challenging to interpret. This makes it difficult to pinpoint the specific data features influencing a suggested action. The lack of interpretability raises concerns, hindering experts from identifying errors and potentially slowing down adoption. Thus, clinicians may be hesitant to embrace recommendations that lack transparent clinical reasoning.

To improve the interpretability and reliability of our \emph{\gls{medt}} model for users, we illustrate the relevance assigned by the transformer to input tokens for an example patient trajectory in Figure~\ref{fig:attention_map}. The relevance across the \gls{atg} components are averaged in Figure~\ref{fig:attention_map}a to better depict the relevance assigned over time, while Figure~\ref{fig:attention_map}b visualizes the importance assigned to each \gls{atg} component. We observe that \emph{\gls{medt}} assigns relatively more importance to time-steps 2 and 3 for this patient sample. Figure Figure~\ref{fig:attention_map}b shows that the model considers the conditioning for the Hepatic of higher relevance in its prediction.

This allows clinicians to monitor the specific points in time when the model assigns the highest importance, facilitating an assessment of its reasonableness. If the model differs from ground-truth clinician actions, an analysis may reveal which features carry the most weight in the decision-making shift. Additionally, if the model relies on clinically irrelevant features, it signals to clinicians that the recommendation may be unsound. This not only enhances understanding of the model's decision process but also invites future research into the reliability of deep \gls{rl} decision-making from a clinical perspective.

\section{Conclusion}
In this work, we propose the \emph{Medical Decision Transformer}, a novel reinforcement learning approach based on the transformer architecture. It models the full context of a patient's medical history to recommend effective sepsis treatment decisions. During training, our framework conditions the model not only on hindsight rewards but also on look-ahead patient acuity scores at each time-step. This enables clinicians to later interact with the model and guide its treatment recommendations by conditioning the model on short-term goals for patient stability. For autoregressive evaluation of our proposed approach, we present a separately trained state predictor that models a patient's clinical state evolution given a sequence of treatment decisions. Our experimental results demonstrate the potential of \emph{\gls{medt}} to bolster clinical decision support systems by providing clinicians with an interpretable and interactive intervention support system.

\section{Acknowledgements}
We are thankful to the Digital Research Alliance of Canada for computing resources and CIFAR, Google, FRQ and NSERC for research funding.

\bibliography{main}
\bibliographystyle{tmlr}

\appendix
\section{Appendix}
\subsection{Acuity Scores and Heuristic}
\label{apd:data}
While many works use SOFA scores for sepsis, recent work shows that both SOFA and SAPS2 predict septic patient mortality effectively~\citep{morkar2022comparative}. We chose SAPS2 as it considers more physiological variables (12) compared to SOFA (6), providing more informative conditioning and allowing flexible user interactions.

Our approach to varying \gls{atg} for \gls{medt} over the sequence involves a simple heuristic: reducing the \gls{atg} linearly at each time-step in the sequence. By varying the \gls{atg} score for individual systems, we found that the hepatic and hematologic systems had the most positive effect on this cohort of patients.

\subsection{Limitations and Future Work}
The MIMIC-III dataset has some limitations, as it only represents a specific geographic area, which could result in an over-representation of certain patient populations and an under-representation of others. Consequently, using the state predictor for evaluation may introduce biases inherent in the dataset on which it was trained. 
To mitigate these potential biases, we will investigate causal representation learning and pre-training techniques that enhance model robustness. Despite these limitations, \emph{\gls{medt}} provides a general framework to harness the vast amount of data found in large-scale \glspl{ehr} from different modalities.
Using the proposed framework, researchers can explore the scalability of the transformer architecture to develop systems for effective treatment recommendation for other medical conditions in the future.

\subsection{Hyperparameters}
\label{apd:hyperparameters}
The transformer architecture of \gls{medt} consists of six layers. The models in this work are trained with batch sizes of 64 and a learning rate of 0.0006. The GELU activation function was used within the transformer architecture.

\subsection{Ablation Study}
From Table \ref{table:ablation}, we observe that \gls{dt} makes a higher performance jump over behavior cloning on patients with negative ground-truth trajectories. This can be attributed to \gls{dt} closely imitating BC for positive trajectories while being able to adopt a more effective policy for negative trajectories due to \gls{rtg} conditioning. On the other hand, \gls{medt} appears to make a higher leap on positive ground-truth patients. This is likely due to the larger number of positive samples available for \gls{medt} to learn and condition using \gls{atg}.

\subsection{Further Visualizations}

\begin{figure*}[ht]
    \centering
    \includegraphics[width=0.8\linewidth]{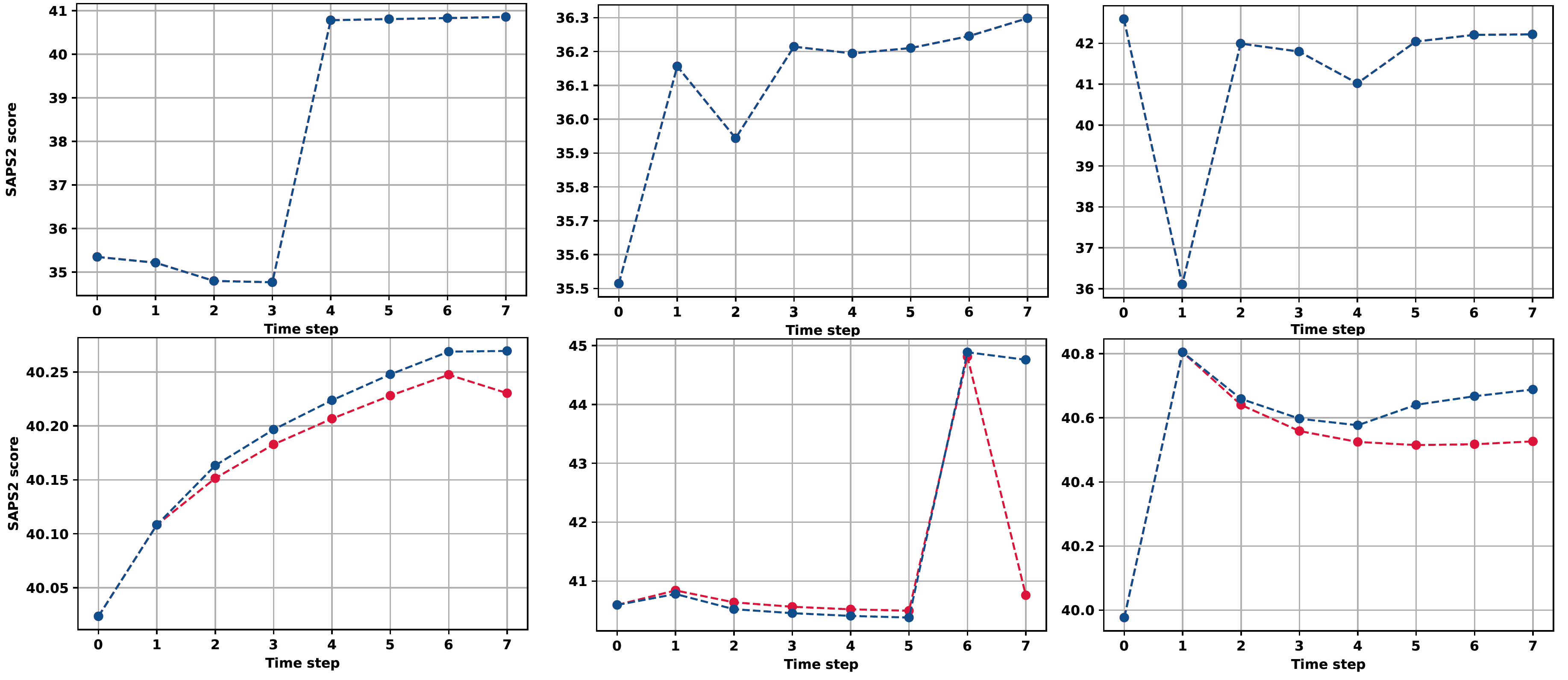}
       
    \caption{Visualization of six patient trajectories computed by the state predictor following treatment recommendation from \gls{dt} (red) and \emph{\gls{medt}} (blue).
}
\label{fig:supp_traj}
\end{figure*}

Figure~\ref{fig:supp_traj} shows the patient trajectories generated by the state predictor, following treatment recommendations from \gls{dt} (red) and \gls{medt} (blue). Specifically, we have visualized the trajectories where the \gls{medt} model produced the same or worse recommendation compared to the \gls{dt} model. 
We postulate that this outcome may be attributable to the limitation of the dataset, which could have resulted in the \gls{medt} model lacking adequate information about the given \gls{atg} conditioning and state pair.

\begin{figure*}[ht]
    \centering
    \includegraphics[width=0.65\linewidth]{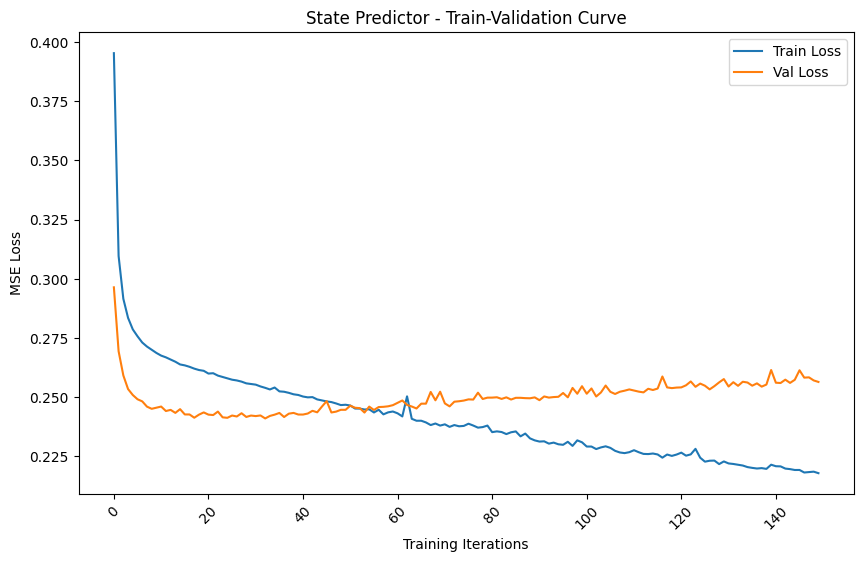}
       
    \caption{Train-validation performance curve of state predictor averaged over 10 seeds.
}
\label{fig:train_val}
\end{figure*}

\begin{figure*}[ht]
    \centering
    \includegraphics[width=0.65\linewidth]{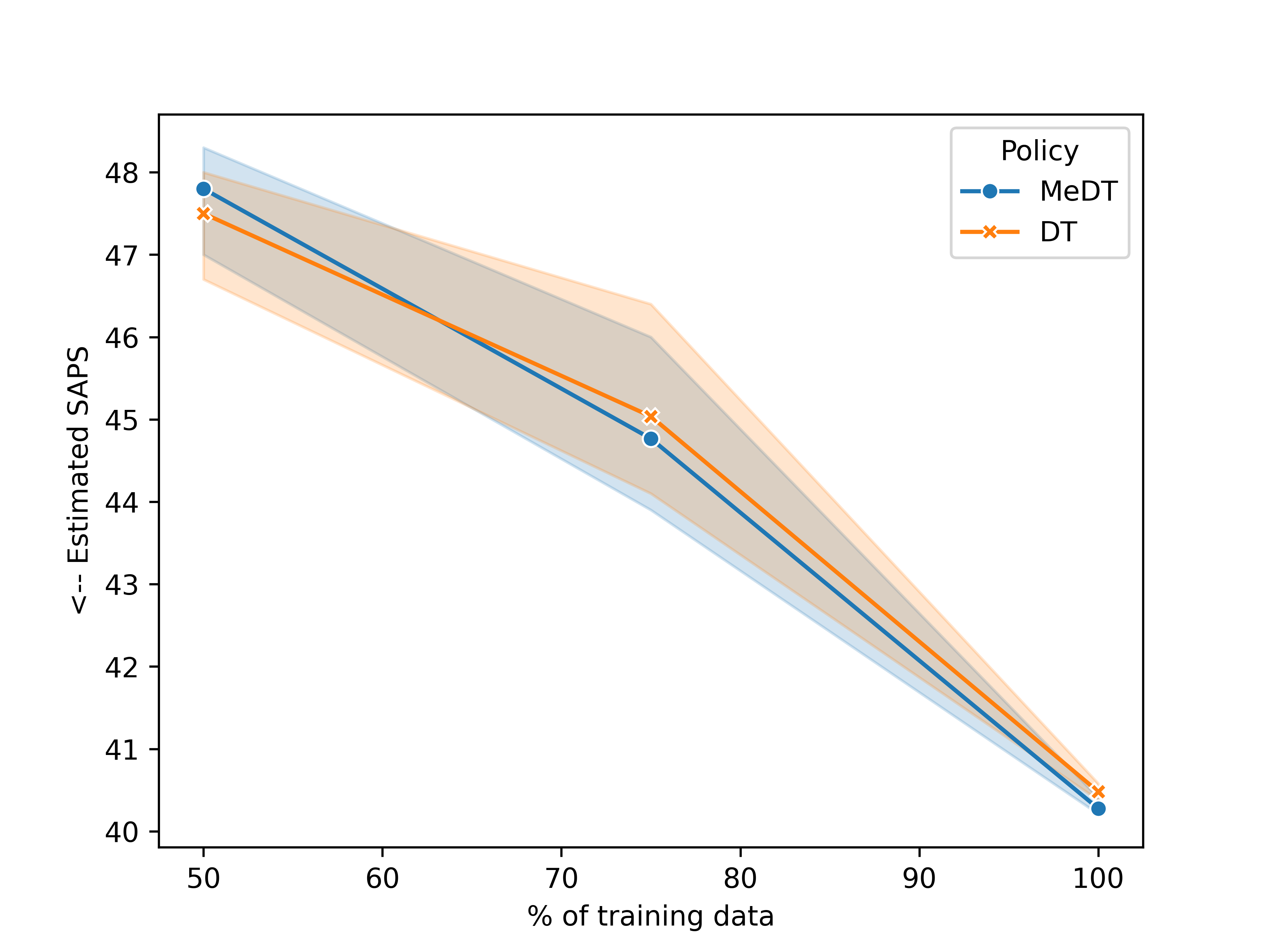}
       
    \caption{Performance of \emph{\gls{medt}} vs \gls{dt} on scaling from training on 50\% to a 100\% of the train split averaged over 3 seeds.}
\label{fig:scale}
\end{figure*}

\begin{table*}[ht]
    \centering
    \caption{List of demographic and observational patient variables in state space.}
    \label{table:obs}
    \begin{tabular}{|ccc|c|}
    \hline & Observations & & Demographics \\
    \hline Glasgow coma scale & Heart rate & Systolic BP & Re-admission \\
    Mean BP & Diastolic BP & Respiratory rate & Gender \\
    Body temp & FiO2 & Potassium & Weight \\
    Sodium & Chloride & Glucose & Ventilation \\
    Magnesium & Calcium & Hemoglobin & Age \\
    WBC & Platelets count & Prothrombin time & \\
    Partial thromboplastin time & Arterial pH & Magnesium & \\
    PaO2 & PaCO2 & Arterial blood gas & \\
    HCO3 & Arterial lactate & PaO2/FiO2 ratio & \\
    SpO2 & SGOT & Creatinine & \\
    Blood urea nitrogen & INR & Bilirubin & \\
    SIRS & Shock index & Cumulative Fluid Balance & \\
    Fluid input/4-hr & Fluid output $/ 4$-hr & Total fluid input & \\
    Total fluid output & & & \\
    \hline
    \end{tabular}
\end{table*}

\begin{table*}[ht]
    \centering
    \caption{Observations considered for the split SAPS2 scores.}
    \label{table:saps}
    \begin{tabular}{|c|c|}
    \hline Organ system & Variable \\
    \hline Cardiovascular $(k c)$ & Blood pressure; Heart rate; Fluid output \\
    Respiratory $(k r)$ & Mechanical ventilation; PaO2 and FiO2 \\
    Neurological $(k n)$ & Glasgow coma scale \\
    Renal $(k l)$ & Blood urea nitrogen \\
    Hepatic $(k h)$ & Bilirubin \\
    Haematologic $(k m)$ & White blood cell count \\
    Other $(k 0)$ & Temperature; Potassium; Sodium \\
    \hline
    \end{tabular}
\end{table*}

\begin{table*}[ht]
    \centering
    \caption{Ablation study on patients with ground-truth positive and negative trajectories.}
    \label{table:ablation}
    \begin{tabular}{|c|c|c|c|}
    \hline Models & All & +1 & -1 \\
    \hline BC & $40.50 \pm 0.03$ & $40.36 \pm 0.02$ & $41.91 \pm 0.09$ \\
    DT & $40.38 \pm 0.03$ & $40.27 \pm 0.02$ & $41.73 \pm 0.09$ \\
    MeDT & $\mathbf{40.31} \pm \mathbf{0.03}$ & $\mathbf{40.15} \pm \mathbf{0.02}$ & $\mathbf{41.72} \pm \mathbf{0.03}$ \\
    \hline
    \end{tabular}
\end{table*}
\end{document}

%% file: math_commands.tex
\usepackage{amsmath,amsfonts,bm}

\def\eqref#1{equation~\ref{#1}}

\def\1{\bm{1}}

\DeclareMathAlphabet{\mathsfit}{\encodingdefault}{\sfdefault}{m}{sl}
\SetMathAlphabet{\mathsfit}{bold}{\encodingdefault}{\sfdefault}{bx}{n}

